\documentclass[]{bytedance_seed}

\usepackage[toc,page,header]{appendix}

\usepackage{amsmath}
\usepackage{minitoc}
\usepackage{subcaption}
\usepackage{graphicx}

\usepackage{amssymb}
\usepackage{dsfont}

\usepackage{mathtools}
\usepackage{amsthm}
\usepackage{multirow}
\usepackage{makecell}
\usepackage{dsfont}
\usepackage{multicol}

\usepackage{algorithm}
\usepackage{algpseudocode}

\usepackage{listings}
\usepackage{courier}
\usepackage{xcolor}
\usepackage{booktabs}
\usepackage{pythonhighlight}

\usepackage{colortbl}
\usepackage{array}
\definecolor{lightblue}{RGB}{220,235,255}
\definecolor{lightgreen}{RGB}{220,255,220}
\definecolor{lightpink}{RGB}{255,220,235}
\definecolor{lightyellow}{RGB}{255,255,200}
\definecolor{lightgray}{RGB}{230,230,230}
\definecolor{Red}{rgb}{1,0,0}

\title{ConceptMoE: Adaptive Token-to-Concept Compression for Implicit Compute Allocation}

\author[*]{Zihao Huang}
\author[*]{Jundong Zhou}
\author[*]{Xingwei Qu}
\author[*]{Qiyang Min}

\author[\dagger]{Ge Zhang}

\affiliation[]{ByteDance Seed}

\contribution[*]{Main authors}
\contribution[\dagger]{Corresponding authors}

\abstract{
Large language models allocate uniform computation across all tokens, ignoring that some sequences are trivially predictable while others require deep reasoning. We introduce ConceptMoE, which dynamically merges semantically similar tokens into concept representations, performing implicit token-level compute allocation. A learnable chunk module identifies optimal boundaries by measuring inter-token similarity, compressing sequences by a target ratio $R$ before they enter the compute-intensive concept model. Crucially, the MoE architecture enables controlled evaluation: we reallocate saved computation to match baseline activated FLOPs (excluding attention map computation) and total parameters, isolating genuine architectural benefits. Under these conditions, ConceptMoE consistently outperforms standard MoE across language and vision-language tasks, achieving +0.9 points on language pretraining, +2.3 points on long context understanding, and +0.6 points on multimodal benchmarks. When converting pretrained MoE during continual training with layer looping, gains reach +5.5 points, demonstrating practical applicability. Beyond performance, ConceptMoE reduces attention computation by up to $R^2\times$ and KV cache by $R\times$. At $R=2$, empirical measurements show prefill speedups reaching 175\% and decoding speedups up to 117\% on long sequences. The minimal architectural modifications enable straightforward integration into existing MoE, demonstrating that adaptive concept-level processing fundamentally improves both effectiveness and efficiency of large language models.

}

\date{\today}
\correspondence{\email{huangzihao.notabot@bytedance.com}}
\checkdata[Code Page]{\url{https://github.com/ZihaoHuang-notabot/ConceptMoE}}

\begin{document}
\maketitle


\section{Introduction}

Large language models (LLMs) process text uniformly at the token level, allocating equal computation to every position in the sequence. Yet not all tokens carry equal semantic weight: while some represent pivotal concepts requiring deep reasoning, others are trivially predictable from context. This one-size-fits-all approach wastes substantial computation on routine predictions while potentially under-resourcing semantically dense content. The natural question emerges: can we move beyond fixed token-level processing toward adaptive concept-level computation?

Traditional approaches to enriching semantic content while reducing token count focus on vocabulary expansion. Recent work~\cite{takase2025large} shows that larger vocabularies consistently improve LLM performance by compressing text into fewer, information-denser tokens. However, the gains are limited: a $100\times$ vocabulary expansion yields only $1.3\times$ compression. The computational cost of massive vocabularies during training and inference makes further scaling impractical.

An alternative paradigm is dynamic token merging within the model. Several studies~\cite{dai2025context,shao2025continuous} merge consecutive tokens into higher-level concepts without vocabulary expansion, but use fixed-length or rule-based strategies that cannot adapt to varying information density. Byte-level transformers~\cite{yu2023megabyte,slagle2024spacebyte,videau2025bytes} explore adaptive chunking~\cite{pagnoni2024byte,hwang2025dynamic}, yet lack precise parameter control and introduce confounding factors through input representation changes. Moreover, most evaluations on dense models compensate reduced token count by scaling depth or width, conflating architectural benefits with parameter increases.

We introduce ConceptMoE, which elevates LLM processing from the token level to the concept level through learnable adaptive chunking. The core insight is straightforward: consecutive tokens with high semantic similarity merge into unified concept representations, while semantically distinct tokens maintain fine-grained granularity. A learnable chunk module identifies optimal boundaries by measuring inter-token similarity, compressing sequences before they enter the compute-intensive concept model. This naturally performs implicit compute allocation - predictable token sequences merge and process efficiently, while complex tokens preserve detailed computation. Crucially, the MoE architecture enables controlled evaluation: by reallocating saved computation to match baseline FLOPs while maintaining identical total parameters, we isolate genuine architectural benefits from concept-level processing.

Our key contributions are:

\begin{itemize}
    \item \textbf{Fair architectural comparison under controlled conditions.} Unlike dense models, MoE allows adjusting activated parameters independently of total parameters. By reallocating computation saved from token reduction, we compare ConceptMoE against standard MoE under identical total parameters and average per-token FLOPs, isolating genuine architectural benefits. We explore three compute reallocation strategies and demonstrate consistent improvements across all configurations.

    \item \textbf{Versatile application across training paradigms.} We demonstrate ConceptMoE's effectiveness across diverse settings: small-scale language model pretraining (Section~\ref{small_exp}), vision-language model training with dual-modality compression (Section~\ref{vlm_exp}), and lossless conversion from pretrained MoE during continual training (Section~\ref{ct_exp}). The CT conversion achieves +5.5 points improvement while enabling direct inference speedup, with from-scratch training reaching +6.4 points overall.
        
    \item \textbf{Inherent efficiency advantages under compute matching.} Even after reallocating saved FLOPs, given compression ratio $R$, ConceptMoE reduces attention map computation by up to $R^2\times$ and KV cache by up to $R\times$ (Table~\ref{table:attn map and kv cache}). Empirical results show prefill speedups up to 175\% and decoding speedups up to 117\% on long sequences under $R=2$, validating both superior performance and efficiency.
    
    \item \textbf{Minimal architectural intrusion for practical adoption.} ConceptMoE requires only a lightweight chunk module and minor decoder modifications (additional QKV projectors in the last 4 layers), making it straightforward to integrate into existing MoE architectures for both pretraining and continual training scenarios.
\end{itemize}

ConceptMoE demonstrates a paradigm shift from uniform token-level to adaptive concept-level processing, fundamentally improving both effectiveness and efficiency of LLMs.
\section{Related Work}
\textbf{Vocabulary size} determines the information capacity of each token in LLMs, typically ranging from 32K to 256K. Larger vocabularies enable higher compression efficiency and fewer tokens per sequence. \citet{tao2024scaling} demonstrate that vocabulary size should scale with model parameters, while \citet{takase2025large} show that increasing vocabulary from 5K to 500K achieves a compression ratio of 1.3, yielding substantial downstream improvements when training on fixed data or token budgets. However, a 100$\times$ vocabulary expansion produces only 1.3$\times$ compression, revealing that further compression gains require exponential vocabulary growth. Moreover, excessively large vocabularies become inference bottlenecks~\cite{cai2024medusa,liu2025csv}. This motivates an alternative approach: dynamic compression within the model itself, which forms the foundation of ConceptMoE.

\textbf{Token-level chunking} compresses tokens internally within the model. Existing approaches include fixed-length merging~\cite{dai2025context,shao2025continuous} and heuristic or rule-based compression~\cite{ankireddytimesqueeze,geng2025zip2zip}. These methods merge multiple tokens into single concepts to achieve higher compression ratios without expanding vocabulary size, but lack adaptive compression strategies. Token information density varies significantly: information-sparse tokens should merge aggressively, while information-rich tokens should maintain finer granularity. Recent concurrent work DLCM\cite{qu2025dynamic} introduces dynamic compression to merge tokens into concepts, but when comparing against FLOPs-matched baselines, the model parameters actually double, undermining the fairness of ablation studies. ConceptMoE addresses these through a learnable chunk module that adaptively identifies optimal merge boundaries based on semantic similarity. Crucially, we leverage MoE properties to ensure all comparisons maintain identical total parameters and average per-token FLOPs, enabling rigorous ablation studies.

\textbf{Byte-level models} require more sophisticated merging strategies due to the granularity of byte tokens, which necessitates aggressive compression to maintain manageable computational costs. AU-Net~\cite{videau2025bytes} introduces multi-level compression but relies on rule-based strategies. BLT~\cite{pagnoni2024byte} leverages a pretrained auxiliary model to compute token entropy, merging low-entropy tokens more aggressively. However, this non-end-to-end approach may encounter difficulties during downstream fine-tuning. H-Net~\cite{hwang2025dynamic} proposes an end-to-end dynamic chunking module, achieving 9$\times$ compression at the byte level. While this represents approximately 2$\times$ compression at the token level, H-Net's experiments control only FLOPs while allowing total parameters to vary, introducing confounding factors. Additionally, the byte-level input representation itself constitutes an experimental variable. ConceptMoE builds on insights from H-Net but operates at the token level with higher compression efficiency and provides fair comparison by controlling both FLOPs and total parameters. We validate effectiveness and efficiency at significantly larger model scales across diverse training scenarios, including language-only pretraining, vision-language training, and continual training from pretrained checkpoints.

\section{Approach}

\subsection{Overview}
\begin{figure}[!ht]
    \centering
    \includegraphics[width=1.0\linewidth]{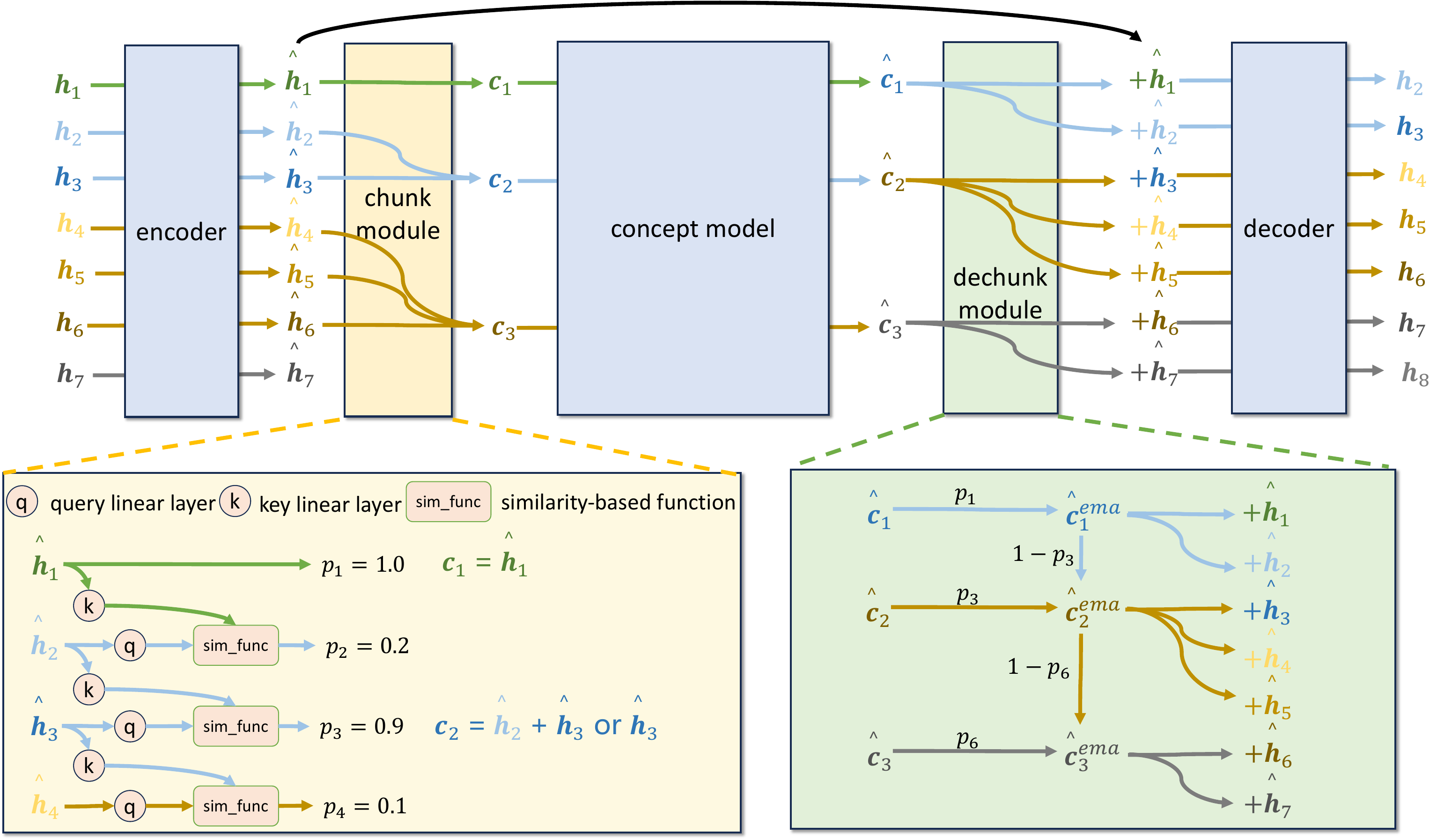}
    \caption{Overview of ConceptMoE, with details of chunk and dechunk modules.}
    \label{fig:overview}
\end{figure}
Just like byte level models, our ConceptMoE is also composed of 5 modules, namely the encoder $\mathcal{E}$, chunk module $\mathsf{Chunk}$, concept model $\mathcal{C}$, dechunk module $\mathsf{DeChunk}$, and decoder $\mathcal{D}$, as shown in Figure~\ref{fig:overview}. $\mathcal{E}$, $\mathcal{C}$ and $\mathcal{D}$ are all composed of multiple stacked MoE modules. Given a sequence of input hidden state $\boldsymbol{H} = \{ \boldsymbol{h}_1, \boldsymbol{h}_2, \dots, \boldsymbol{h}_n,\dots, \boldsymbol{h}_N \}$, $\boldsymbol{h}_n \in \mathbb{R}^d$,

\begin{equation}
\begin{aligned}
\hat{\boldsymbol{H}} = \mathcal{E}(\boldsymbol{H}), \quad  %
\boldsymbol{C},\boldsymbol{P}  = \mathsf{Chunk}(\hat{\boldsymbol{H}}), \\
\hat{\boldsymbol{C}}  = \mathcal{C}(\boldsymbol{C}), \quad  %
\boldsymbol{Z}  = \mathsf{DeChunk}(\hat{\boldsymbol{C}}, \boldsymbol{P}), \\
\hat{\boldsymbol{Z}} = \mathcal{D}(\hat{\boldsymbol{C}},\boldsymbol{Z}),  \quad\quad\quad
\end{aligned}
\end{equation}
where $\boldsymbol{C} = \{ \boldsymbol{c}_1, \boldsymbol{c}_2, \dots, \boldsymbol{c}_m,\dots, \boldsymbol{c}_M \}$, $\boldsymbol{c}_m \in \mathbb{R}^d$\footnote{For the sake of simplicity, the dimensions of $\boldsymbol{c}$ and $\boldsymbol{h}$ are kept consistent here, both being $d$; in fact, the dimension of $\boldsymbol{c}$ can be larger than that of $\boldsymbol{h}$.} is the concept embeddings, $M\le N$, $\hat{\boldsymbol{H}}$, $\boldsymbol{Z}$, $\hat{\boldsymbol{Z}}$ and $\boldsymbol{H}$ have the same dimension. $\boldsymbol{P} = \{ p_1, p_2, \dots,p_n,\dots,p_N \}$ is the probability that the token is a chunk boundary. Note that in the $\mathcal{D}$ input, each token is guaranteed to have an associated concept for joint decoding. This significantly enhances model capability by ensuring that information in the concept is fully utilized across multiple subsequent tokens.

In general, the computational proportion of $\mathcal{E}$ and $\mathcal{D}$ is relatively small, and the main FLOPs come from the $\mathcal{C}$. The $\mathsf{Chunk}$ will select to merge multiple consecutive $\hat{\boldsymbol{h}}$ into single $\boldsymbol{c}$ to enter $\mathcal{C}$. We regard the merged multiple tokens as a concept, hence the intermediate model is called the concept model. We provide PyTorch-like code in the Appendix~\ref{apdx:code} for better understanding.

\subsection{Chunk module} \label{sec: chunk module}
The chunk module aims to identify optimal chunk boundaries for a given input token sequence under a specified compression ratio. Similar to H-Net\cite{hwang2025dynamic}, it selects which consecutive tokens are easily predictable and merges them into chunks, while keeping hard-to-predict tokens unmerged or merged into smaller chunks. This approach performs implicit \textbf{token-level compute allocation}, in contrast to explicit compute allocation methods like MoE (e.g., Zero Expert\cite{jin2024moe++}) that activate different amounts of computation for each token.

Specifically, for the input $\boldsymbol{h}_n$, we calculate the cosine similarity between the embeddings of two adjacent tokens after linear transformation to determine if this token serves as a chunk boundary:

\begin{equation}
\boldsymbol{q_n} = W_q\boldsymbol{h_n}, \quad \boldsymbol{k_n} = W_k\boldsymbol{h_n}, \quad p_n = \frac{1}{2}(1-\frac{\boldsymbol{q_n}^\top \boldsymbol{k_{n-1}}}{\parallel \boldsymbol{q_n} \parallel \cdot  \parallel \boldsymbol{k_{n-1}} \parallel }),\quad  b_n=\mathds{1}_{p_n\ge 0.5},
\end{equation}
where $W_q, W_k \in \mathds{R}^{d\times d}$ are learnable parameters, and we set $p_1=1.0$ to ensure that the first token in the decoder $\mathcal{D}$ always has a concept for joint decoding. If the probability $p_n$ exceeds 0.5, the token is identified as a chunk boundary. This design is natural and intuitive: when several consecutive tokens exhibit high similarity but the current token shows low similarity to its predecessor, it indicates a significant semantic shift. Such tokens typically carry substantially different information and require more careful processing. 

\textbf{Auxiliary loss} Given a target compression ratio, we introduce an auxiliary loss to constrain the token compression ratio on the training set, where the compression ratio is defined as $R=N/M \ge1$. Inspired by the load balancing loss in MoE, we treat boundary and non-boundary selections as two experts and constrain their activation frequencies to achieve the target compression ratio. Let the average probabilities and average selection ratios for boundary and non-boundary be:
\begin{equation}
G_1 = \frac{1}{N}\sum_{n}^{N}p_n, \quad   G_2 = \frac{1}{N}\sum_{n}^{N}1-p_n=1-G_1, \quad   
F_1 = \frac{1}{N}\sum_{n}^{N}b_n, \quad F_2 = \frac{1}{N}\sum_{n}^{N}1-b_n=1-F_1,
\end{equation}
then the auxiliary loss is
\begin{equation}
\mathcal{L} _{aux}=RF_1G_1+\frac{R}{R-1}F_2G_2= \frac{R}{R-1}((R-1)F_1G_1 + (1-F_1)(1-G_1)).
\end{equation}

Note that when computing $G_1$ and $F_1$, we aggregate statistics across all samples in the current devices rather than averaging per-sample statistics. This enables \textbf{sample-level compute allocation}: when a batch contains both difficult and easy samples, the model can reduce the compression ratio for difficult samples while increasing it for easy samples. In practical use, we use $\lambda$ to control the weight of this auxiliary loss.

\textbf{Random flip boundary} When we constrain the compression ratio on the training set, empirical observations show that distribution shifts in the evaluation set can lead to excessively high compression ratio, which speeds up inference but significantly degrades model performance. We address this by adding random perturbations to boundaries during training to simulate such scenarios, thereby mitigating the over-compression issue caused by distribution shifts at inference time. Specifically, after computing $p_n$, we sharpen the probability:
\begin{equation}
p_n^{sharp} = \left\{\begin{matrix}p_n^{\frac{1}{\tau } }
  & p_n\ge 0.5\\
  1 -(1-p_n)^{\frac{1}{\tau } }& p_n<0.5
\end{matrix}\right.
\end{equation}
Where $\tau$ is a hyperparameter to control sharpness. The final boundary is obtained by sampling from a Bernoulli distribution parameterized by $p_n^{sharp}$:
\begin{equation}
b_n^{train} \sim \text{Bernoulli}(p_n^{sharp}), \quad b_n^{train} \in \{0,1\}, \quad \Pr(b_n^{train}=k) = 
\begin{cases} 
p_n^{sharp} & \text{if } k=1 \\
1-p_n^{sharp} & \text{if } k=0
\end{cases}
\end{equation}
The sharpened probability distribution is more likely to flip boundaries with low confidence (close to 0.5) during sampling, while probabilities with high confidence (close to 0 or 1) are less likely to be flipped. This ensures stable training convergence.

\textbf{Merging strategy}  We can either sum all tokens within a chunk to obtain the concept, which maximally preserves the information of each token, or use only the last token as the concept, since the self-attention mechanism in the encoder $\mathcal{E}$ enables the last token to already aggregate the information of the entire chunk.

\subsection{Dechunk module}
The dechunk module remaps concepts back to tokens and ensures that no information leakage occurs. Before dechunk, we first apply exponential moving average (EMA) on concepts. Let  $\mathcal{I} = \{n \mid b_n = 1, 1 \leq n \leq N\}$ with $ |\mathcal{I}| = M$. Define the index mapping

\begin{equation}
\phi: \{1, 2, \ldots, m,\ldots, M\} \to \mathcal{I}, \quad \phi(m) = n_m,
\end{equation}
then for each $\hat{c}_m$, there is
\begin{equation}
\hat{\boldsymbol{c}}_m^{ema} = p_{\phi(m)}\hat{\boldsymbol{c}}_m + (1-p_{\phi(m)})\hat{\boldsymbol{c}}_{m-1}.
\end{equation}
This mechanism is illustrated in the green part of Figure~\ref{fig:overview}. The EMA accelerates chunking convergence through the following process: consider an example where "Simple and easy-to-" and "understand picture" are initially split into two concepts, with $p=0.5$ for "understand picture". Through EMA, if the model discovers that the concept of "Simple and easy-to-" effectively aids in predicting tokens within "understand picture", it will reduce the boundary probability $p$ for "understand picture". Once $p$ falls below 0.5, the boundary is eliminated, merging "Simple and easy-to-" and "understand picture" into a unified concept.

After EMA, based on the given index mapping, dechunk the concept to the token-level embedding. Specifically, we define the mapping from concept index to token index:
\begin{equation}
\psi: \{1, 2, \ldots, n,\ldots, N\} \to \{1, 2, \ldots, m\ldots, M\}, \quad \psi(n) = m \text{ if } \phi(m) \leq n < \phi(m+1).
\end{equation}
Finally we can obtain the $\mathcal{D}$ input $z_n$ as
\begin{equation}
\boldsymbol{z_n}=\hat{\boldsymbol{h}}_n + \hat{\boldsymbol{c}}_{\psi(n)}^{ema}.
\end{equation}

\subsection{Joint decoding} \label{sec: joint decoding}
During token decoding, multiple tokens share the same concept in the dechunking process. To fully exploit the information in the concept, which accounts for most of the model's computation and contains rich information, we perform joint decoding of $\boldsymbol{z_n}$ and $\hat{\boldsymbol{c}}_{\psi(n)}^{ema}$. Specifically, in each self-attention layer of the decoder $\mathcal{D}$, we have:
\begin{equation}
\text{Attention}(\boldsymbol{z_n},\hat{\boldsymbol{c}}_{\psi(n)}^{ema}) = \text{softmax}\left(\frac{(\boldsymbol{z_n}W_q+{\color{Red} \hat{\boldsymbol{c}}_{\psi(n)}^{ema}W_q^c} )(\boldsymbol{z_n}W_k+{\color{Red} \hat{\boldsymbol{c}}_{\psi(n)}^{ema}W_k^c} )^T}{\sqrt{d_{head}}} + M\right)(\boldsymbol{z_n}W_v+{\color{Red} \hat{\boldsymbol{c}}_{\psi(n)}^{ema}W_v^c} ),
\end{equation}
where $M$ is the causal attention mask with $M_{ij} = -\infty \cdot \mathds{1}_{i < j}$. The highlighted terms show how we augment standard attention by incorporating concept information into the query, key, and value. Given the shallow depth of decoder $\mathcal{D}$, the additional parameters are negligible while yielding substantial performance gains. Moreover, this design maintains compatibility with existing architectures, enabling straightforward application to continue training (CT) from pretrained models.

\subsection{Compute reallocation strategies for fair comparison} \label{sec:Compute_reallocation}
Given a standard MoE model, we decompose it into encoder, concept model, and decoder with layer depths $L_{\mathcal{E}}$, $L_{\mathcal{C}}$, and $L_{\mathcal{D}}$ respectively. Temporarily disregarding attention map computation, let $C_{attn}$ and $C_{moe}$ denote the FLOPs per token in self-attention and MoE layers. The concept model incurs $L_{\mathcal{C}}(C_{attn}+C_{moe})$ FLOPs per token. With the chunk module at compression ratio $R$, the encoder outputs $R$ tokens on average before the concept model processes one token, reducing its computation to $\frac{L_{\mathcal{C}}(C_{attn}+C_{moe})}{R}$(an $R$-fold reduction). This freed computation can be reallocated by increasing $L_{\mathcal{C}}$, $C_{attn}$, or $C_{moe}$ to maintain total FLOPs. Unlike dense architectures, MoE allows adjusting activated parameters while fixing total parameters. This property enables a fair comparison: \textbf{under identical total parameters and per-token FLOPs, we can isolate the true gains of ConceptMoE over standard MoE by varying only the activated parameters.} This paper explores three parameter allocation schemes:
\begin{enumerate}
    \item \textbf{Increasing $C_{moe}$}: We match the compute by increasing the number of activated experts in the MoE layers. This is the simplest approach and can be applied to both pretraining and continual training.
    \item \textbf{Increasing $L_{\mathcal{C}}$ and $C_{moe}$}: Building on the previous approach, we additionally increase $L_{\mathcal{C}}$ by looping through intermediate layers. This method is CT-friendly and introduces no additional parameters.
    \item \textbf{Increasing $C_{attn}$ and $C_{moe}$}: We scale up both self-attention and MoE computation while keeping total parameters fixed. This is achieved by proportionally enlarging the hidden size of $L_{\mathcal{C}}$ while reducing the total number of MoE experts. This requires two additional linear projectors for the mappings $\hat{\boldsymbol{h}} \to \boldsymbol{c}$ and $\hat{\boldsymbol{c}} \to \boldsymbol{z}$. This approach is less suitable for CT and better suited for pretraining.
\end{enumerate}

We now quantify the reduction in attention map computation and kv cache for the three schemes, which represents an inherent advantage of ConceptMoE. In standard MoE, the attention map computation for the concept model is $L_{\mathcal{C}}dN^2$, where $d$ denotes the hidden size and $N$ the sequence length. Assume using multi-head attention, the kv cache is $2L_{\mathcal{C}}dN$. The corresponding computations and kv cache for the three schemes are shown in Table~\ref{table:attn map and kv cache}, where $L_{loop}$ is the number of loop layers. Any of these schemes can significantly reduce the computation of the attention map and the kv cache.

\begin{table}[h]
\centering
\caption{Attention map computation and KV cache comparison}
\begin{tabular}{lcccc}
\toprule
\textbf{Method} & \textbf{Attn Map FLOPs} & \textbf{Reduction} & \textbf{KV Cache} & \textbf{Reduction} \\
\midrule
Baseline (MoE) & $L_{\mathcal{C}}dN^2$ & $1\times$ & $2L_{\mathcal{C}}dN$ & $1\times$ \\
Increasing $C_{moe}$ & $\frac{L_{\mathcal{C}}dN^2}{R^2}$ & $R^2\times$ & $\frac{2L_{\mathcal{C}}dN}{R}$ & $R\times$ \\
Increasing $L_{\mathcal{C}}$ and $C_{moe}$ & $\frac{(L_{\mathcal{C}}+L_{loop})dN^2}{R^2}$ & $\frac{R^2L_{\mathcal{C}}}{L_{\mathcal{C}}+L_{loop}}\times$ & $\frac{2(L_{\mathcal{C}}+L_{loop})dN}{R}$ & $\frac{RL_{\mathcal{C}}}{L_{\mathcal{C}}+L_{loop}}\times$ \\
Increasing $C_{attn}$ and $C_{moe}$ & $\frac{L_{\mathcal{C}}dN^2}{R^{1.5}}$ & $R^{1.5}\times$ & $\frac{2L_{\mathcal{C}}dN}{\sqrt{R}}$ & $\sqrt{R}\times$ \\
\bottomrule
\end{tabular}
\label{table:attn map and kv cache}
\end{table}
\section{Experiments}
We conduct comprehensive experiments to validate ConceptMoE's effectiveness and efficiency across diverse settings. Our evaluation encompasses four main aspects: (1) Small-scale language model pretraining (Section~\ref{small_exp}) at 12B and 24B parameters to establish core benefits under controlled conditions. (2) Vision-language model training (Section~\ref{vlm_exp}) at 60B parameters, demonstrating dual-modality compression and achieving strong gains on long context tasks. (3) Continual training conversion (Section~\ref{ct_exp}) from pretrained MoE at 90B parameters, validating practical deployment through lossless integration. (4) Inference speedup analysis (Section~\ref{sec:speedup}) on 300B parameters models, measuring actual latency improvements across diverse compression ratios and layer configurations. Additionally, we perform extensive ablation studies examining auxiliary loss weight, chunking strategies, router design, joint decoding, boundary noise, and target compression ratios. All experiments maintain identical total parameters and per-token FLOPs (excluding attention maps) between ConceptMoE and MoE, ensuring fair architectural comparison.

\textbf{Common model configuration} The MoE baseline activates 8 experts. ConceptMoE uses an auxiliary loss weight of $\lambda=0.03$. For token merging strategy, models with CT integration use only the last token in each chunk as the concept to minimize structural modifications when converting MoE to ConceptMoE and keep the initial loss as low as possible. Other models sum tokens within each chunk to form concepts. The hyperparameter for boundary random flipping is set to $\tau=6$, under which approximately 4\% of tokens are flipped. This configuration keeps the evaluation compression ratio close to the training compression ratio without affecting training performance.

\textbf{Evaluation Benchmarks} We conduct extensive evaluations on both open-source and proprietary benchmarks. The evaluation suite covers text reasoning, mathematics, code generation, knowledge retrieval, needle-in-haystack, long context summarization and understanding, as well as multimodal tasks including visual localization, visual reasoning, hallucination detection, visual question answering, and chart extraction. Detailed benchmark specifications are provided in the Appendix~\ref{apdx:eval}.

\subsection{Small-scale language model pretraining} \label{small_exp}
\textbf{Model configurations} We evaluate ConceptMoE on two MoE configurations: a 0.5B FLOPs model with 12B total parameters (MoE-A0.5B-12B), and a 1B FLOPs model with 24B total parameters(MoE-A1B-24B). Both baselines activate 8 experts per token. For ConceptMoE, we set $L_{\mathcal{E}}=L_{\mathcal{D}}=4$ and scale the hidden size of $\mathcal{C}$ to $4/3$ that of $\mathcal{E}$ and $\mathcal{D}$. This configuration increases the per-token compute of $\mathcal{C}$ by a factor of $16/9$ relative to the baseline. We therefore set the compression ratio to $R=16/9$ to match the total compute budget \footnote{Note that all FLOPs comparisons exclude attention map computation, meaning ConceptMoE actually consumes fewer FLOPs than the baseline when compute-matched.}. This setup corresponds to the third reallocation strategy in Section~\ref{sec:Compute_reallocation}, which jointly increases $C_{attn}$ and $C_{moe}$.

\textbf{Training Protocol} We define tokens per parameter (TPP) as the ratio of training tokens to total model parameters. All models are trained at TPP=400: MoE-A0.5B-12B on 243B tokens and MoE-A1B-24B on 559B tokens. We use the AdamW optimizer with cosine learning rate decay.

\textbf{Results} Table~\ref{tab:model_comparison} presents the main results comparing ConceptMoE against standard MoE baselines. Both model pairs maintain identical total parameters and per-token FLOPs, differing only in compute allocation strategy under concept or token level process. ConceptMoE consistently outperforms the baseline across most metrics. 

These results validate that concept-level processing provides genuine benefits beyond simple token-level processing, suggesting that adaptive chunking effectively allocates more compute to semantically complex token sequences while efficiently processing predictable patterns.

\begin{table}[h]
\centering
\caption{Model performance comparison. Comp.: Comprehensive Evaluation, Reason.: Reasoning, Know.: Knowledge. Details see Table~\ref{tab:obm easy}.}
\small
\begin{tabular}{lcccccccccc}
\toprule
\multirow{2}{*}{\textbf{Model}} & \multicolumn{2}{c}{\textbf{Training}} & \textbf{Eval} & \multicolumn{6}{c}{\textbf{Openbench Easy}} \\
\cmidrule(lr){2-3} \cmidrule(lr){4-4} \cmidrule(lr){5-10}
& \textbf{Tokens} & \textbf{Loss}$\downarrow$ & \textbf{Loss}$\downarrow$ & \textbf{Comp.}$\uparrow$ & \textbf{Reason.}$\uparrow$ & \textbf{Math}$\uparrow$ & \textbf{Code}$\uparrow$ & \textbf{Know.}$\uparrow$ & \textbf{All}$\uparrow$ \\
\midrule
MoE-A0.5B-12B & 243B & 1.852 & 1.992 & 46.2 & 37.6 & 27.8 & 30.7 & \textbf{26.2} & 35.6 \\
ConceptMoE-A0.5B-12B & 243B & \textbf{1.849} & \textbf{1.990} & \textbf{47.3} & \textbf{39.1} & \textbf{28.8} & 30.7 & 26.1 & \textbf{36.4} \\
\midrule
MoE-A1B-24B & 559B & 1.717 & 1.851 & \textbf{57.6} & 54.4 & 47.2 & 42.3 & \textbf{41.6} & 50.0 \\
ConceptMoE-A1B-24B & 559B & \textbf{1.711} & \textbf{1.844} & 57.4 & \textbf{56.8} & \textbf{50.0} & \textbf{42.4} & 41.5 & \textbf{50.9} \\
\bottomrule
\end{tabular}
\label{tab:model_comparison}
\end{table}

\subsection{Train a vision-language model}\label{vlm_exp}

\textbf{Model configurations} We investigate the potential of concept representations as inputs for vision-language model (VLM). We conduct experiments on MoE-A2.5B-60B and ConceptMoE-A2.5B-60B. The vision encoder uses a small ViT\cite{dehghani2023patch} for image feature extraction, and a linear projection to align the dimensions of the visual token with the text tokens. On the LLM side, we maintain $L_{\mathcal{E}}=L_{\mathcal{D}}=4$ and apply the third reallocation strategy from Section~\ref{sec:Compute_reallocation}. We scale the hidden size of $\mathcal{C}$ to 1.5× while slightly reducing the MoE inner dimension, and set the compression ratio to $R=2$ to preserve total activated compute. Note that in the VLM setting, we apply compression to both visual and textual tokens.

\textbf{Training Protocol} We first pretrain the LLM on 500B tokens, then integrate a pretrained NaViT and a randomly initialized linear projection layer. The VLM continues training for 200B tokens on a mixture of high-quality image-text pairs and pure text data with 32K sequence length. Training uses the AdamW optimizer with cosine learning rate decay throughout.

\begin{figure}[h]
    \centering
    \includegraphics[width=1\linewidth]{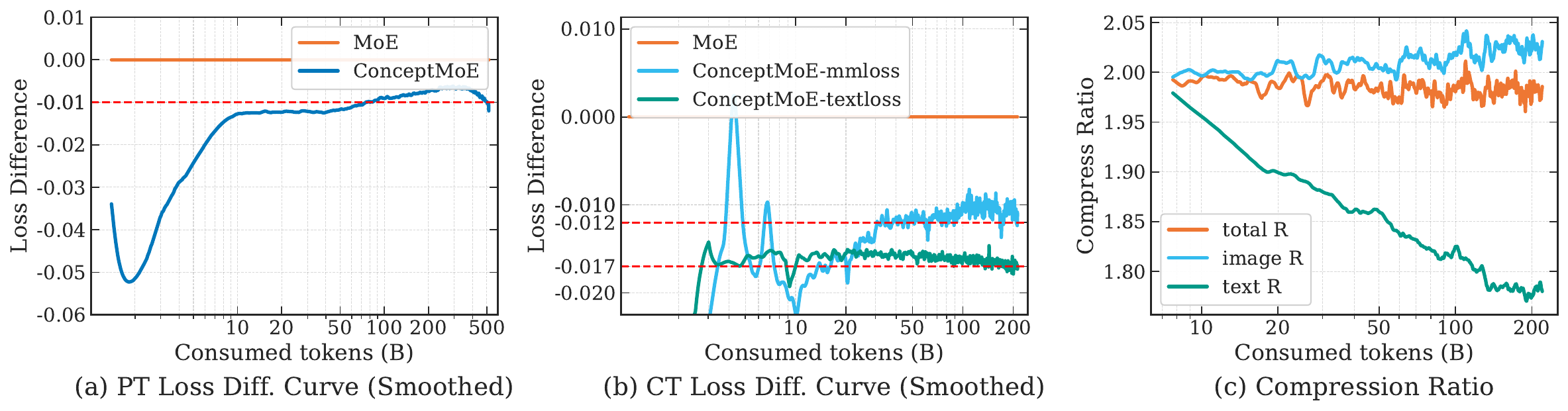}
    \caption{Training dynamics of loss and compression ratio. (a) Loss difference between ConceptMoE and MoE during language model pretraining (PT). (b) Loss difference during multimodal continue training(CT), separated into image-text data (mmloss) and text-only data(textloss). (c) Compression ratio evolution for image tokens and text tokens during multimodal training.}
    \label{fig:vlm_loss}
\end{figure}

\textbf{Results} Figure~\ref{fig:vlm_loss} shows training dynamics. During PT (a), ConceptMoE achieves 0.01 lower loss than MoE. In multimodal CT (b), this gap increases to 0.017 for text data and 0.012 for image-text data. This difference reflects adaptive compression (c): while maintaining overall $R=2$, the model compresses text less (using more compute) and images more, suggesting higher visual redundancy.

Tables~\ref{tab:text_benchmark}, \ref{tab:long_context}, and \ref{tab:vlm_benchmark} present downstream results with average compression ratios of 2.27, 2.0, and 2.0 respectively. ConceptMoE outperforms MoE by 0.9 points on text benchmarks, 2.3 on long context, and 0.6 on multimodal tasks. Long context shows particularly strong gains across most subtasks, as expected from reduced sequence length alleviating degradation on long documents. The Needle task improvement confirms that concept merging preserves information. Multimodal results show gains in reasoning and understanding, validating stronger concept-level reasoning. However, fine-grained visual tasks (location, chart text, image Q\&A) decline slightly, likely because treating image tokens sequentially disrupts spatial relationships critical for localization.

\begin{table}[h]
\centering
\caption{Text benchmark results on OpenBench. Comp.: Comprehensive Evaluation, Spec.: Specialized Knowledge, Know.: Knowledge. Details see Table~\ref{tab:obm all}.}
\small
\begin{tabular}{lcccccccc}
\toprule
\textbf{Model} & \textbf{Comp.}$\uparrow$ & \textbf{Reasoning}$\uparrow$ & \textbf{Math}$\uparrow$ & \textbf{Code}$\uparrow$ & \textbf{Spec.}$\uparrow$ & \textbf{Know.}$\uparrow$ & \textbf{All}$\uparrow$ \\
\midrule
MoE-A2.5B-60B & 41.7 & 35.0 & 16.3 & 34.6 & 28.8 & 35.0 & 33.5 \\
ConceptMoE-A2.5B-60B & \textbf{41.8} & \textbf{35.3} & 16.2 & \textbf{36.9} & \textbf{30.8} & \textbf{36.8} & \textbf{34.4} \\
\bottomrule
\end{tabular}
\label{tab:text_benchmark}
\end{table}

\begin{table}[h]
\centering
\caption{Long context evaluation results. LC: Long context, Sum.: Summary, Underst: Understanding.}
\small
\begin{tabular}{lcccccc}
\toprule
\textbf{Model} & \textbf{Needle}$\uparrow$ & \textbf{LC Learn.}$\uparrow$ & \textbf{LC Reason.}$\uparrow$ & \textbf{LC Sum./Q\&A}$\uparrow$ & \textbf{LC Underst.}$\uparrow$ & \textbf{All}$\uparrow$ \\
\midrule
MoE-A2.5B-60B & 78.9 & 34.7 & 10.3 & \textbf{63.7} & 40.9 & 49.4 \\
ConceptMoE-A2.5B-60B & \textbf{80.7} & \textbf{38.8} & \textbf{17.1} & 59.0 & \textbf{45.1} & \textbf{51.7} \\
\bottomrule
\end{tabular}
\label{tab:long_context}
\end{table}

\begin{table}[h]
\centering
\caption{Vision-language benchmark results. Reason.: Reasoning, Desc. Halluc.: Description Hallucinations, Q\&A: Image Question and Answer, Comp.: Comprehensive Bench.}
\small
\begin{tabular}{lccccccc}
\toprule
\textbf{Model} & \textbf{Location}$\uparrow$ & \textbf{Reason.}$\uparrow$ & \textbf{Desc. Halluc.}$\uparrow$ & \textbf{Q\&A}$\uparrow$ & \textbf{Chart Text}$\uparrow$ & \textbf{Comp. }$\uparrow$ & \textbf{All}$\uparrow$ \\
\midrule
MoE-A2.5B-60B & \textbf{82.2} & 53.5 & \textbf{35.8} & \textbf{61.4} & \textbf{34.7} & 53.9 & 53.6 \\
ConceptMoE-A2.5B-60B & 81.9 & \textbf{57.9} & 35.3 & 58.0 & 33.6 & \textbf{58.7} & \textbf{54.2} \\
\bottomrule
\end{tabular}
\label{tab:vlm_benchmark}
\end{table}

\subsection{Train from CT} \label{ct_exp}
In this subsection, we investigate converting MoE to ConceptMoE during continual training and evaluate the performance gap compared to training from scratch. We observe three key findings:

\begin{itemize}
    \item \textbf{CT conversion is lossless and highly beneficial.} ConceptMoE-top15 maintains baseline performance, while adding layer loops yields substantial gains of +5.5 points on Open Benchmark. Training from scratch achieves an additional +0.9 points, reaching +6.4 overall improvement.

    \item \textbf{PT evaluation shows distribution shift effects.} During PT, downstream metrics may appear lower due to misalignment between evaluation and pretraining data distributions, resulting in higher compression ratios ($R=1.81$). However, evaluation and CT data distributions align closely, stabilizing compression at the expected ratio ($R=1.5$).

    \item \textbf{CT-converted models enable direct inference speedup.} The stabilized compression ratio of $R=1.5$ in CT allows lossless conversion of existing MoE models to ConceptMoE while obtaining significant inference gains. Section~\ref{sec:speedup} presents detailed speedup analysis showing prefill acceleration up to 43.6\% and decoding acceleration up to 53.3\% at $R=1.5$.
\end{itemize}

\begin{figure}[ht]
    \centering
    \includegraphics[width=1\linewidth]{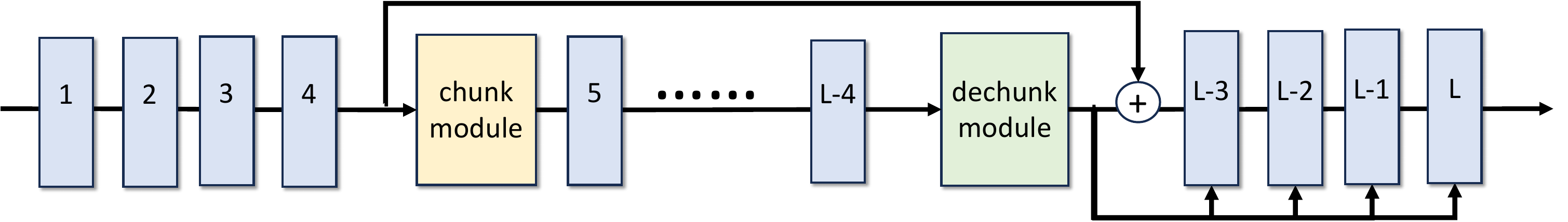}
    \caption{From MoE to ConceptMoE illustration. The blue blocks denote the original MoE components. We add a chunk module and a dechunk module. In addition, in the last four self-attention layers, we insert an extra QKV projector initialized to zeros to enable joint decoding of concepts and tokens.}
    \label{fig:ct}
\end{figure}

As illustrated in Figure~\ref{fig:ct}, we start with MoE-A2.5B-90B pretrained on 700B tokens and convert it to ConceptMoE by adding a chunk module (query and key linear layers with random initialization), a dechunk module, and additional QKV projectors in the last 4 self-attention layers (zero-initialized). We then perform 400B tokens of 32k CT, followed by 40B tokens of 128k CT, and finally 3B tokens of in-house SFT with a small fraction of math and reasoning problems accompanied by long chain-of-thought. Given the CT setting, we conservatively set $R=1.5$ and explore two compute allocation schemes:
\begin{itemize}
    \item ConceptMoE-top15\footnote{We omit activated parameters and total parameters for brevity.}: Increase activated experts from 8 to 15 (Strategy 1 in Section~\ref{sec:Compute_reallocation}).
    \item ConceptMoE-top11-loop8: Increase activated experts from 8 to 11 while looping 8 intermediate layers (Strategy 2 in Section~\ref{sec:Compute_reallocation}).
\end{itemize}
These configurations ensure nearly identical average FLOPs and total parameters, enabling fair comparison. Additionally, we train ConceptMoE-top11-loop8 from scratch to quantify the performance gap introduced by CT conversion.

\begin{figure}[h]
    \centering
    \includegraphics[width=1\linewidth]{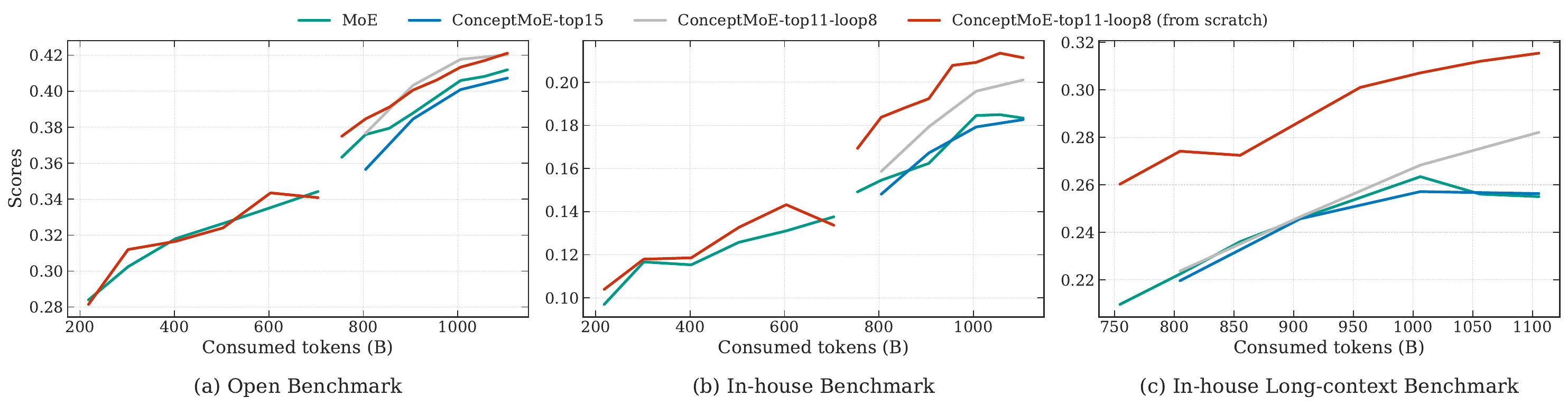}
    \caption{Evolution of evaluation metrics across three benchmarks throughout the training process.}
    \label{fig:ct_training}
\end{figure}

Figure~\ref{fig:ct_training} shows downstream benchmark evolution during training. During CT (800B to 1100B tokens), ConceptMoE-top15 matches MoE performance closely, with only a 0.3 point gap on Open Benchmark. ConceptMoE-top11-loop8 shows clear improvements, indicating that reallocating compute to $L_{\mathcal{C}}$ is more efficient than increasing $C_{moe}$. Comparing CT conversion to from-scratch training, ConceptMoE-top11-loop8 (from scratch) performs similarly on Open Benchmark but shows substantial gains on in-house and long context benchmarks, particularly the latter. This highlights the value of from-scratch training while demonstrating that CT conversion preserves and moderately improves existing model capabilities. Evaluation compression ratios remain stable at $R=1.5$ throughout CT.

During PT (0 to 700B tokens), ConceptMoE-top11-loop8 and MoE exhibit comparable fluctuations, with evaluation compression ratio at $R=1.81$, reducing FLOPs by 12.5\%. This discrepancy between PT ($R=1.81$) and CT ($R=1.5$) compression ratios reveals distribution shift: evaluation and CT data are more aligned, both representing higher quality than PT data.

Table~\ref{tab:ct_results} presents post-SFT performance. ConceptMoE-top15 achieves a modest 0.4 point improvement over MoE, while ConceptMoE-top11-loop8 shows substantial gains of 5.5 points overall, with particularly strong improvements in reasoning (+8.3), math (+12.2), and code (+6.4). While layer looping partly contributes to these gains, ConceptMoE critically achieves them without increasing FLOPs. The from-scratch trained model further improves by 0.9 points over the CT-converted version, demonstrating significant benefits. Evaluation compression ratios stabilize at $R=1.5$ across all configurations.

\begin{table}[h]
\centering
\footnotesize
\caption{CT results comparison on Open Benchmark. Comp. Eval.: Comprehensive Evaluation, FS: From Scratch.}
\begin{tabular}{lcccc}
\toprule
\textbf{Category} & \textbf{MoE} & \textbf{ConceptMoE-top15} & \textbf{ConceptMoE-top11-loop8} & \textbf{ConceptMoE-top11-loop8(FS)} \\
\midrule
Overall & 40.9 & 41.3 & 46.4 & \textbf{47.3} \\
\midrule
Comp. Eval. & 49.4 & 50.5 & \textbf{54.4} & 53.9 \\
Reasoning & 30.3 & 28.4 & \textbf{38.6} & 38.5 \\
Math & 38.1 & 39.7 & 50.3 & \textbf{52.8} \\
Code & 20.0 & 20.9 & 26.4 & \textbf{30.1} \\
Instruction & \textbf{54.7} & 52.9 & \textbf{55.1} & 54.6 \\
Knowledge & \textbf{28.2} & 26.2 & 27.5 & 27.3 \\
Multilingual & 71.7 & 75.3 & 75.3 & \textbf{80.3} \\
\bottomrule
\end{tabular}
\label{tab:ct_results}
\end{table}

\subsection{Significant inference speedup} \label{sec:speedup}
We evaluated the inference latency of ConceptMoE on Hopper GPUs. Using MoE-A10B-300B as the baseline, we assessed 5 ConceptMoE configurations, ranging from efficiency-oriented to quality-oriented setups. Our results show that quality-oriented ConceptMoE achieves \textbf{comparable speed} to MoE on short sequences despite \textbf{doubling the number of layers}, while maintaining increasing speedup advantages on long sequences. Efficiency-oriented ConceptMoE delivers speedups of \textbf{32.1\%} to \textbf{117.1\%} in decoding and \textbf{24.7\%} to \textbf{175.1\%} in prefill.

\begin{figure}[h]
    \centering
    \includegraphics[width=1\linewidth]{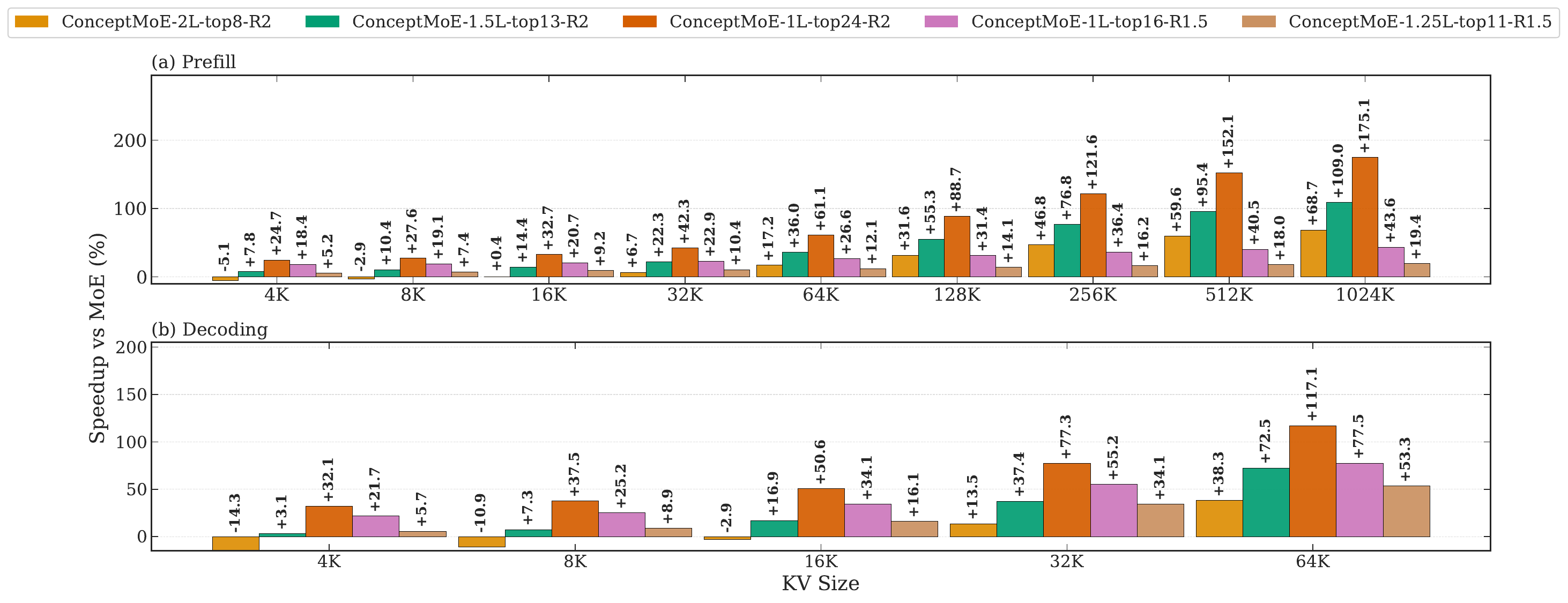}
    \caption{Inference latency speedup over MoE for prefill and decoding. The prefill plot uses sequence length on the x axis, and the decoding plot uses KV cache length on the x axis with batch size 256. The y axis reports speedup relative to MoE in percent. ConceptMoE-$x$L-top$y$-R$z$ matches MoE in FLOPs and total parameters, where $x$ is the layer multiplier, $y$ is the number of activated experts per MoE block with baseline 8 and increased to match FLOPs, and $z$ is the compression ratio.}
    \label{fig:speedup_rate}
\end{figure}

Specifically, we configure models in the form ConceptMoE-$x$L-top$y$-R$z$, where $x$ represents the layer multiplication ratio, $y$ denotes the number of activated experts (baseline is 8), and $z$ indicates the compression ratio. For quality-oriented configurations, we use ConceptMoE-2L-top8-R2. For efficiency-oriented setups, we employ ConceptMoE-1L-top24-R2 and ConceptMoE-1L-top16-R1.5. For balanced configurations, we adopt ConceptMoE-1.5L-top13-R2 and ConceptMoE-1.25L-top11-R1.5. The $R=1.5$ configurations are similar to the models discussed in Section~\ref{ct_exp}.

\begin{figure}[h]
    \centering
    \includegraphics[width=0.9\linewidth]{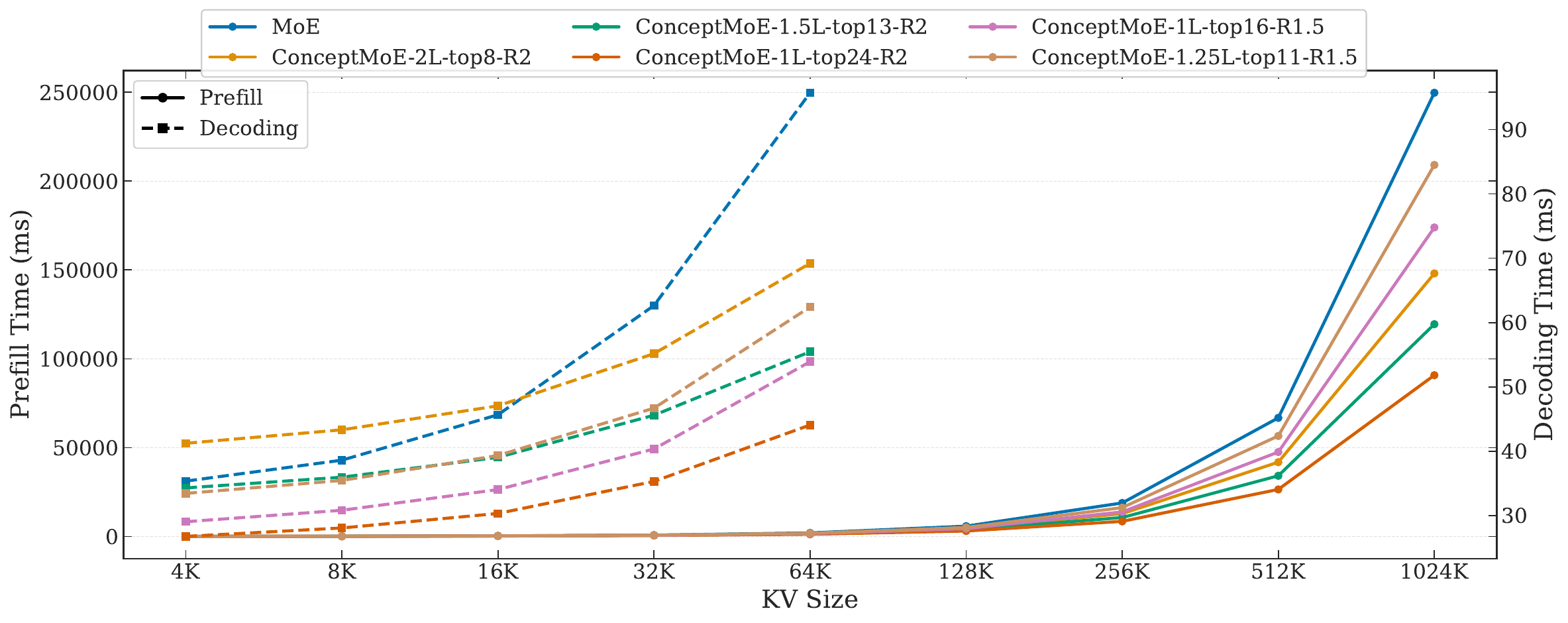}
    \caption{Inference latency for prefill and decoding. The prefill plot uses sequence length on the x axis, and the decoding plot uses KV cache length on the x axis with batch size 256. The y axis reports the measured end to end inference latency on Hopper GPUs. ConceptMoE-$x$L-top$y$-R$z$ matches MoE in FLOPs and total parameters, where $x$ is the layer multiplier, $y$ is the number of activated experts per MoE block with baseline 8 and increased to match FLOPs, and $z$ is the compression ratio.
        }
    \label{fig:latency}
\end{figure}

Figure~\ref{fig:speedup_rate} shows the speedup ratio of ConceptMoE over MoE. For the prefill stage, we evaluate speedup variations across input sequences ranging from 4K to 1024K. For the decoding stage, we assess speedup changes for kv cache lengths from 4K to 64K at batch size 256. Figure~\ref{fig:latency} presents the actual latency. We observe that even when doubling the number of layers at $R=2$, ConceptMoE still achieves substantial speedup—this stems from the quadratic reduction in attention map computation and the linear reduction in kv cache. When keeping the layer count unchanged, efficiency-oriented configurations achieve prefill speedups up to \textbf{175\%}, which continue to increase with longer sequences, and decoding speedups up to \textbf{117.1\%}. At $R=1.5$, we also observe prefill speedups up to \textbf{43.6\%} and decoding speedups up to \textbf{53.3\%}. Section~\ref{ct_exp} validates the feasibility of integrating ConceptMoE in CT at $R=1.5$, which allows us to directly convert existing MoE models to ConceptMoE losslessly while obtaining significant inference gains.

\subsection{Structure ablation}

\subsubsection{Auxiliary loss weight}
We ablate the auxiliary loss weight $\lambda$ on ConceptMoE-A0.5B-12B trained for 243B tokens with target compression ratio $R=2$. Following H-Net\cite{hwang2025dynamic}'s auxiliary loss weight of 0.03, we evaluate $\lambda \in \{0.03, 0.1, 0.5, 1.0\}$ and examine their effects on training loss and achieved compression ratio.

\begin{figure}[h]
    \centering
    \includegraphics[width=0.9\linewidth]{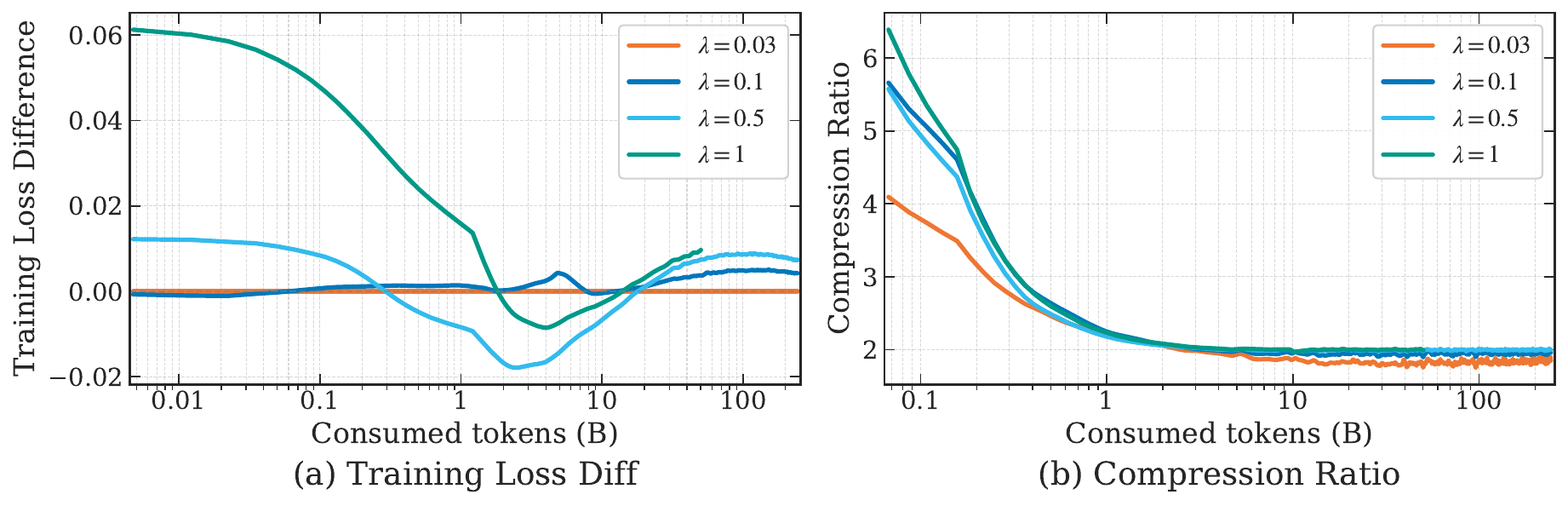}
    \caption{Impact of auxiliary loss weight $\lambda$ on training loss and compression ratio.}    \label{fig:aux_loss_weight}
\end{figure}

Figure~\ref{fig:aux_loss_weight} presents the results. As $\lambda$ increases, training loss degrades while all configurations achieve compression ratios close to 2. Based on these findings, we set $\lambda=0.03$ for all other experiments.

\subsubsection{Chunking strategy}
We compare two chunking strategies on ConceptMoE-A0.5B-12B trained for 243B tokens with target compression ratio $R=2$: (1) Dynamic Chunk: our learnable adaptive chunking based on token similarity, (2) Fixed Chunk: merging every consecutive $R$ tokens into one concept. We use No Chunk(MoE-A0.5B-12B) as our baseline.

\begin{figure}
    \centering
    \includegraphics[width=1\linewidth]{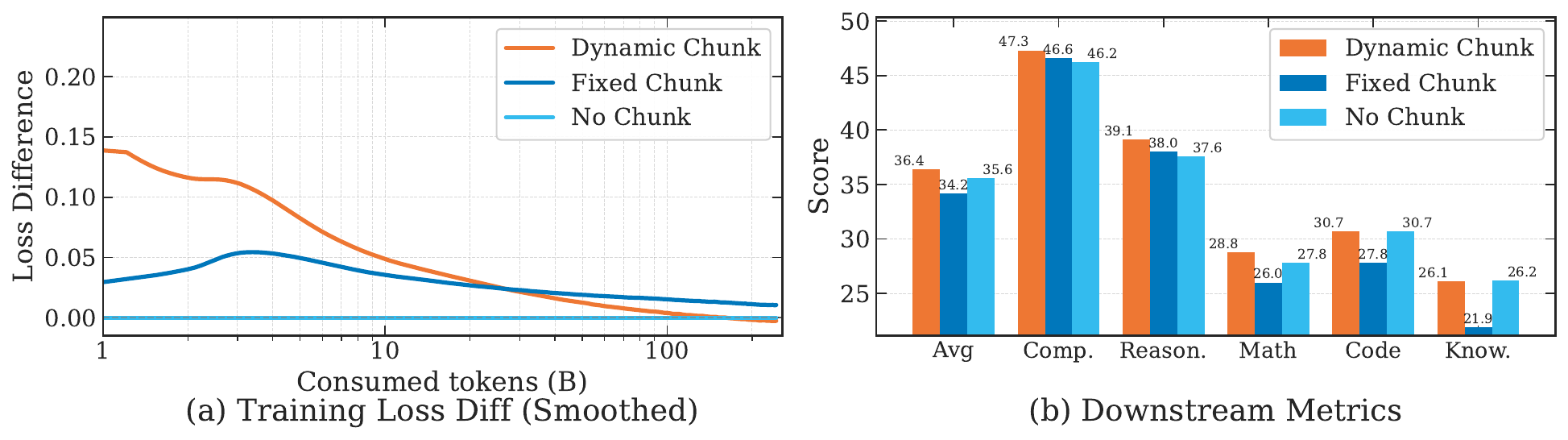}
    \caption{Comparison of chunking strategies. (a) Training loss difference relative to No Chunk baseline (i.e. MoE) during pretraining. Dynamic Chunk consistently achieves lower loss than Fixed Chunk. (b) Downstream benchmark scores after training. Comp.: Comprehensive Evaluation, Reason.: Reasoning, Know.: Knowledge.}
    \label{fig:chunk_strategy}
\end{figure}

Figure~\ref{fig:chunk_strategy}(a) shows training dynamics. Dynamic Chunk maintains a consistent advantage throughout training and ultimately achieves 0.004 lower loss than No Chunk, demonstrating that adaptive compression with proper compute reallocation improves optimization. In contrast, Fixed Chunk degrades by 0.01 relative to No Chunk, indicating that uniform merging disrupts the model's learning dynamics. Downstream evaluation (Figure~\ref{fig:chunk_strategy}(b)) confirms this pattern: Dynamic Chunk achieves the highest average score (36.4), outperforming No Chunk (35.6) and Fixed Chunk (34.2). These results validate that adaptive boundary identification preserves semantic coherence while enabling efficient computation.

\subsubsection{Router design}

Section~\ref{sec: chunk module} introduces our cosine similarity-based score for identifying chunk boundaries. A natural alternative is to directly predict boundary scores using a linear layer, analogous to MoE routers. We compare these two designs: cosine router (our approach) and linear router, on ConceptMoE-A0.5B-12B trained for 243B tokens with target compression ratio $R=2$.

\begin{figure}
    \centering
    \includegraphics[width=1\linewidth]{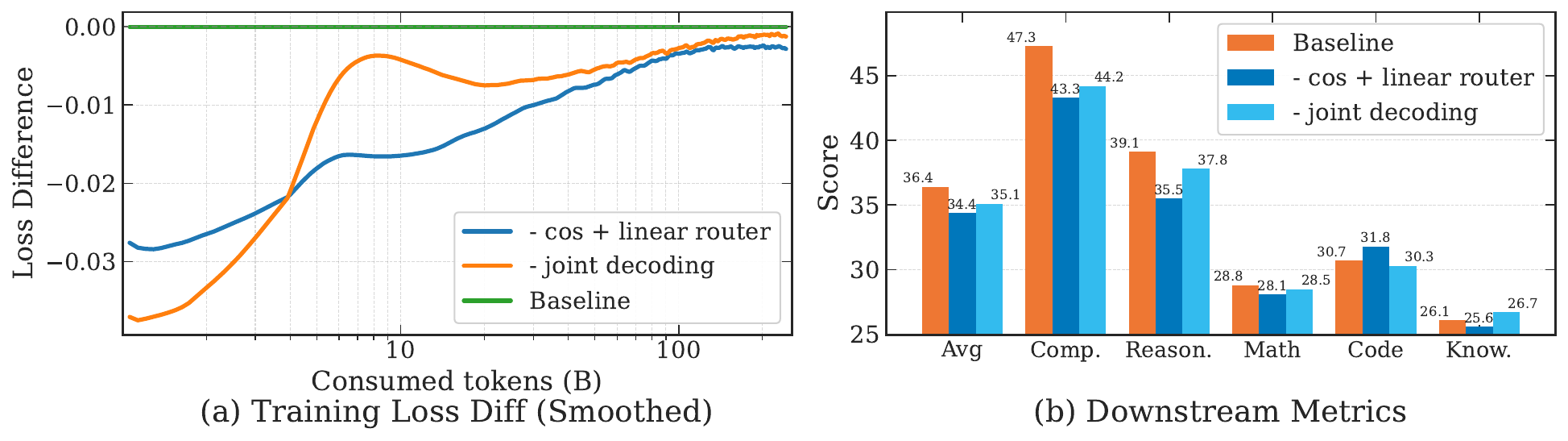}
    \caption{Impact of router type and joint decoding on training and downstream performance. }
    \label{fig:router and joint decoding}
\end{figure}

Figure~\ref{fig:router and joint decoding}(a) shows that the linear router achieves 0.003 lower training loss at convergence. However, Figure~\ref{fig:router and joint decoding}(b) reveals a substantial gap in downstream performance: the linear router scores 34.4 average, significantly underperforming the cosine router at 36.4. This train-eval discrepancy suggests that the linear router overfits to training data patterns. In contrast, the cosine router's explicit modeling of inter-token similarity provides better generalization by capturing semantic relationships rather than memorizing dataset-specific boundary patterns.

\subsubsection{Joint decoding ablation}

Section~\ref{sec: joint decoding} introduces joint decoding of concepts and tokens in the decoder's final layers, adding negligible computation. We ablate this component on ConceptMoE-A0.5B-12B trained for 243B tokens with target compression ratio $R=2$.

Figure~\ref{fig:router and joint decoding}(a) shows that removing joint decoding yields 0.002 lower training loss at convergence. However, Figure~\ref{fig:router and joint decoding}(b) reveals substantial downstream degradation: the model without joint decoding scores 35.1 average compared to 36.4 with joint decoding. This pattern mirrors the router ablation: improvements in training loss do not guarantee better generalization.

We hypothesize that joint decoding acts as implicit regularization during training. By forcing the decoder to explicitly attend to concept information through additional QKV projections, the model learns more robust representations that transfer better to downstream tasks. Without joint decoding, the decoder may overfit to residual token-level patterns in the training data while underutilizing the semantic information encoded in concepts. This validates that leveraging concept representations throughout the decoding process is essential for realizing the full benefits of adaptive compression.

\subsection{Boundary Noise for Robustness}

We observe a discrepancy between training and evaluation compression ratios: the model achieves lower compression during training than during evaluation. Analysis of the boundary probability distribution reveals that a substantial portion of probabilities cluster around 0.5. During evaluation, these borderline cases can easily flip, causing unintended compression ratio drift.

Theoretically, for target compression ratio $R$, the mean probability should satisfy $\mathbb{E}[p_n] \approx 1/R$. However, without noise, we observe significantly lower probability means, indicating that many boundaries hover near the 0.5 threshold. During evaluation, slight perturbations or distribution shifts push these marginal probabilities over the boundary, leading to higher compression than intended.

To improve robustness, we introduce noise during training to simulate evaluation conditions. We evaluate Bernoulli noise with $\tau \in \{4, 6\}$ for $p_n^{sharp}$ and Gaussian noise with $\sigma=0.1$. All models train with target $R=1.5$.

\begin{figure}[h]
    \centering
    \includegraphics[width=1\linewidth]{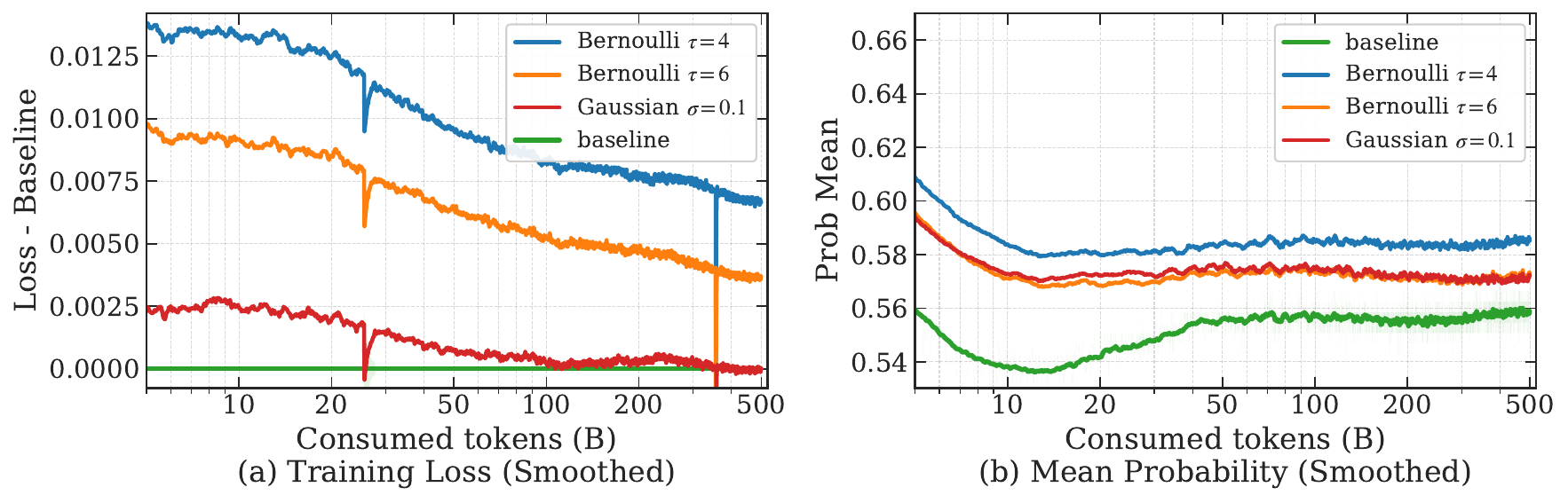}
    \caption{Training loss diff and mean probability of chunk module for different noise strategies. We do smoothing and remove some spike data for better visualization, which doesn't affect the conclusion.}
    \label{fig:loss_prob_mean_noise_ablation}
\end{figure}

Figure~\ref{fig:loss_prob_mean_noise_ablation} shows training dynamics. Bernoulli noise strategies yield higher training loss at convergence, with smaller $\tau$ producing larger degradation, confirming that noise perturbs chunk boundaries. Gaussian noise maintains comparable training loss. Critically, noise regularization normalizes the probability mean closer to $1/R$ (e.g. 0.667), stabilizing evaluation compression ratios.

Table~\ref{tab:noise_ablation_results} demonstrates that noise improves downstream performance despite higher training loss. ConceptMoE with $\tau=4$ achieves 30.3 average score (+1.4 over baseline), with stronger Bernoulli noise proving most effective. This validates that boundary noise strengthens chunk module robustness by preventing probability collapse around 0.5, ensuring consistent compression behavior between training and evaluation. To avoid excessive impact on training loss, we use Bernoulli noise with $\tau=6$ in all other experiments.

\begin{table}[h]
\centering
\footnotesize
\caption{Model performance comparison on OpenBench for different noise strategies. Comp.: Comprehensive Evaluation, Spec.: Specialized Knowledge, Know.: Knowledge.. Comp.: Comprehensive Evaluation, Reason.: Reasoning, Know.: Knowledge.}
\begin{tabular}{lcccc}
\toprule
\textbf{Category} & \textbf{ConceptMoE} & \textbf{ConceptMoE-Gaussian} & \textbf{ConceptMoE-$\tau=4$} & \textbf{ConceptMoE-$\tau=6$} \\
\midrule
All   & 28.9 & 29.6 & \textbf{30.3} & 30.0 \\
\midrule
Comp. Eval. & 38.9 & 40.0 & 40.1 & \textbf{40.3} \\
Reasoning       & 22.9 & 23.5 & \textbf{24.7} & 24.0 \\
Math       &  9.1 &  8.7 &  9.4 &  \textbf{9.5} \\
Code       & 36.0 & 37.5 & \textbf{38.8} & 38.7 \\
Spec.   & 25.0 & \textbf{28.4} & 27.6 & 26.1 \\
Know.       & 30.2 & 28.6 & \textbf{30.3} & 29.6 \\
\bottomrule
\end{tabular}
\label{tab:noise_ablation_results}
\end{table}

\subsubsection{Target compression ratio}

We investigate the impact of compression ratio by training models with $R=2$ and $R=4$, using MoE-A1B-24B as baseline with compute reallocation strategy 3 (increasing $C_{attn}$ and $C_{moe}$ from Section~\ref{sec:Compute_reallocation}) over 559B tokens.

\begin{figure}[h]
    \centering
    \includegraphics[width=1\linewidth]{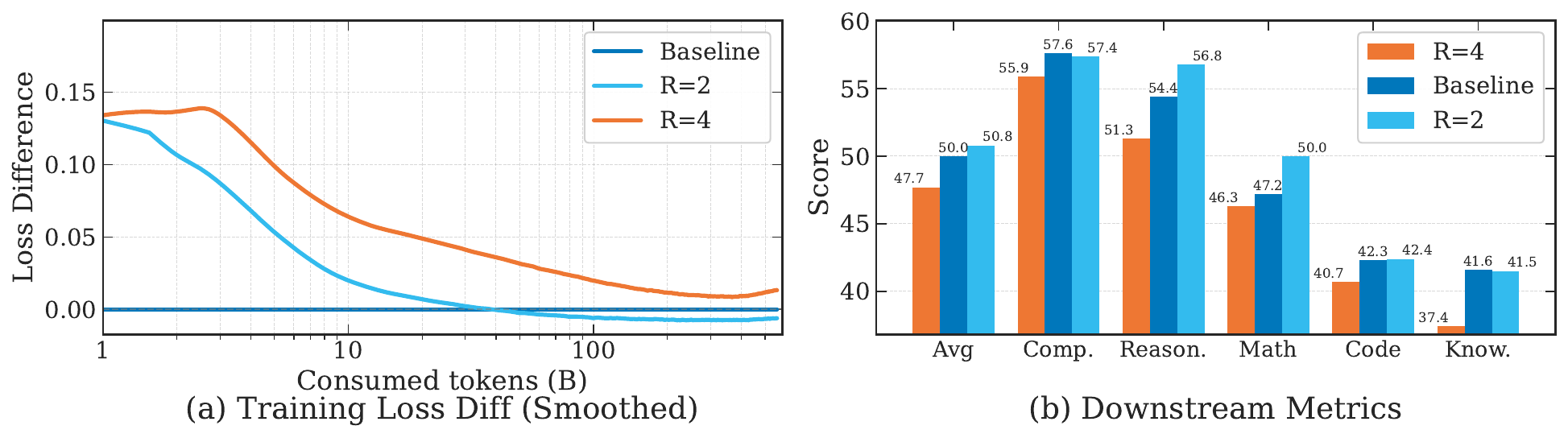}
    \caption{Impact of target compression ratio on training and downstream performance. (a) Training loss difference relative to baseline. $R=2$ converges close to baseline, while $R=4$ shows persistent degradation. (b) Downstream benchmark scores. $R=2$ outperforms the baseline across most metrics, while $R=4$ underperforms substantially, particularly on reasoning and math tasks. Results indicate that excessive compression (e.g., $R=4$) degrades performance despite successful compression ratio control.}    
    \label{fig:R4}
\end{figure}

Figure~\ref{fig:R4} reveals that higher compression does not guarantee better performance. While $R=2$ achieves comparable training loss to the baseline and improves downstream scores to 50.8 average, $R=4$ shows substantial degradation in both training loss (0.013 gap at convergence) and downstream performance (47.7 average). The gap is particularly pronounced on reasoning (51.3 vs 56.8) and math (46.3 vs 50.0), suggesting that aggressive compression disrupts complex reasoning patterns.

We hypothesize that each dataset has an optimal compression ratio determined by its semantic redundancy distribution. At $R=4$, the model is forced to merge tokens with significant semantic differences, losing critical information for downstream tasks. The auxiliary loss successfully constrains the compression ratio during training, but the resulting 4$\times$ compression fundamentally exceeds the natural redundancy level in the data. This indicates that compression ratio should be calibrated to dataset characteristics rather than maximized unconditionally. For typical pretraining corpora, $R=1.5$ to $R=2$ appears to strike an effective balance between efficiency and information preservation.

\section{Conclusion}

We introduce ConceptMoE, a framework that dynamically merges semantically similar tokens into concepts through learnable chunking. By operating within the MoE architecture, we enable fair comparison under controlled conditions: reallocating saved computation to match baseline FLOPs and total parameters isolates genuine architectural benefits. Experiments demonstrate consistent improvements across language pretraining (+0.9 points), vision-language training(+0.6 points), long context (+2.3 points), and continual training conversion (+5.5 points with layer looping).

Beyond performance gains, ConceptMoE reduces attention computation by up to $R^2\times$ and KV cache by $R\times$, achieving prefill speedups up to 175\% and decoding speedups up to 117\% at $R=2$. The minimal architectural changes enable straightforward integration into existing systems. ConceptMoE demonstrates that adaptive concept-level processing fundamentally improves both the effectiveness and efficiency of LLMs, opening new directions for compute allocation in sparse architectures.

\clearpage

\bibliographystyle{plainnat}
\bibliography{main}

\clearpage

\beginappendix

\section{Evaluation benchmark} \label{apdx:eval}

We provide the evaluation datasets included in each category of OpenBench. Table~\ref{tab:obm easy} details the relatively easy evaluation sets used for models with 12B and 24B total parameters. Table~\ref{tab:obm all} includes both easy and hard evaluation sets used for models with 60B and 90B total parameters.

\begin{table}[htbp]
\centering
\caption{Open benchmarks  Easy across different domains}
\begin{tabular}{@{}llllll@{}}
\toprule
Comprehensive Evaluation & Reasoning & Math & Code & Knowledge \\ \midrule
MMLU\cite{hendryckstest2021,hendrycks2021ethics}        & BBH\cite{suzgun2022challenging}  & MATH\cite{hendrycksmath2021} & HumanEval\cite{chen2021codex} & TriviaQA\cite{2017arXivtriviaqa} \\
C-Eval\cite{huang2023ceval}      & DROP\cite{Dua2019DROP} &      & MBPP+\cite{austin2021program}     & ChineseSimpleQA\cite{he2024chinesesimpleqachinesefactuality} \\
MMLU-Pro\cite{wang2024mmlu}    &      &      & McEval\cite{mceval}    & \\
AGIEval\cite{zhong2023agieval}     &      &      &           & \\ \bottomrule
\end{tabular}
\label{tab:obm easy}
\end{table}

\begin{table}[htbp]
\centering
\small
\caption{Open benchmarks across different domains}
\begin{tabular}{@{}lllllll@{}}
\toprule
\thead{Comprehensive\\Evaluation} &
\thead{Reasoning} &
\thead{Math} &
\thead{Code} &
\thead{Knowledge} &
\thead{Specialized\\Knowledge} \\ \midrule
MMLU\cite{hendryckstest2021,hendrycks2021ethics}        & BBH\cite{suzgun2022challenging}        & MATH\cite{hendrycksmath2021}        & HumanEval\cite{chen2021codex}      & TriviaQA\cite{2017arXivtriviaqa}        & GPQA\cite{rein2024gpqa} \\
C-Eval\cite{huang2023ceval}      & DROP\cite{Dua2019DROP}       & AIME2024    & MBPP+\cite{austin2021program}          & SimpleQA\cite{wei2024measuring}        & \\
MMLU-Pro\cite{wang2024mmlu}    & ARC\_AGI\cite{chollet2019measure}   & AIME2025    & LiveCodeBench\cite{jain2024livecodebench}  & ChineseSimpleQA\cite{he2024chinesesimpleqachinesefactuality} & \\
AGIEval\cite{zhong2023agieval}     & ProcBench\cite{fujisawa2024procbench}  & HARP\cite{yue2024harp}        & McEval\cite{mceval}         &                 & \\
HLE\cite{phan2025humanity} & ZebraLogic\cite{zebralogic2024} & Omni-MATH\cite{gao2024omnimathuniversalolympiadlevel} & & & \\
SuperGPQA\cite{pteam2025supergpqascalingllmevaluation}   & KOR-Bench\cite{ma2024korbenchbenchmarkinglanguagemodels}  & OlympiadBench\cite{he2024olympiadbench} & & & \\
LiveBench\cite{livebench}   &            &             & & & \\ \bottomrule
\end{tabular}
\label{tab:obm all}
\end{table}

\section{Code} \label{apdx:code}
Below we present the forward function in PyTorch format. In this example, the encoder has 2 layers, the concept model has 23 layers, and the decoder has 2 layers. The hidden size is 2048. The input consists of 1024 tokens, which are compressed to 701 tokens before entering the concept model.

\begin{python}
import torch
import torch.nn.functional as F
from torch import nn

class ChunkModule(nn.Module):
    def __init__(self, config):
        super().__init__()
        self.q_proj_layer = nn.Linear(config.hidden_size, config.hidden_size, bias=False)
        self.k_proj_layer = nn.Linear(config.hidden_size, config.hidden_size, bias=False)
    
    def forward(self, hidden_states):
        cos_sim = torch.einsum(
                    "l d, l d -> l",
                    F.normalize(self.q_proj_layer(hidden_states[:, :-1]), dim=-1),
                    F.normalize(self.k_proj_layer(hidden_states[:, 1:]), dim=-1),
                            )       # shape [1023,]
        boundary_prob = torch.clamp(((1 - cos_sim) / 2), min=0.0, max=1.0)  # shape [1023,]
        # Force boundary probability of the first element to 1.0
        PAD_PROB = 1.0
        boundary_prob = F.pad(boundary_prob, (1, 0), "constant", PAD_PROB) # shape [1024,]

        selected_idx = torch.zeros_like(boundary_prob, dtype=torch.long)
        boundary_mask = boundary_prob >= 0.5
        selected_idx[..., boundary_mask] = 1
        boundary_prob = torch.stack(((1 - boundary_prob), boundary_prob), dim=-1)

        selected_probs = boundary_prob.gather(
                    dim=-1, index=selected_idx.unsqueeze(-1)
                )  # (shape hidden_states.shape[:-1], 1)
        return boundary_prob, boundary_mask, selected_probs

class DechunkModule(nn.Module):
    def __init__(self, config):
        super().__init__()

    def forward(self, concept, boundary_prob, boundary_mask):
        concept_prob = boundary_prob[boundary_mask]     # shape [701,]
        concept_merge = torch.zeros_like(concept)

        # For ease of understanding, this is written in for-loop form. In practice, it can be accelerated through parallel scan.
        concept_merge[0] = concept[0]                   # shape [701, 2048]
        for i in range(1, concept.shape[0]):
            concept_merge[i] = concept_merge[i-1]*(1-concept_prob[i]) + concept[i] * concept_prob[i]

        plug_back_idx = boundary_mask.cumsum(dim=0) - 1
        concept_merge = torch.gather(
                    concept_merge, dim=0, index=plug_back_idx.expand(-1, 2048)
                )            # concept_merge shape [701,2048] -> [1024,2048]
        return concept_merge
                
class STE(torch.autograd.Function):
    @staticmethod
    def forward(ctx, x):
        return torch.ones_like(x)
    @staticmethod
    def backward(ctx, grad_output):
        grad_x = grad_output
        return grad_x
def ste_func(x):
    return STE.apply(x)

class ConceptMoE(nn.Module):
    def __init__(self, config):
        super().__init__()
        self.encoder = nn.Modulelist(transformer_layer(layer_id=i) for i in range(2))
        self.concept_model = nn.Modulelist(transformer_layer(layer_id=i) for i in range(2,25))
        self.decoder = nn.Modulelist(transformer_layer(layer_id=i) for i in range(25,27))

        self.lm_head = nn.Linear(config.hidden_size, config.vocab_size, bias=False)
        self.embedding = nn.Embedding(config.vocab_size, config.hidden_size)

        self.chunk_module = ChunkModule(config)
        self.dechunk_module = DechunkModule(config)

    def forward(self, input_ids):
        # encoder
        hidden_state = self.embedding(input_ids)        # shape [1024, 2048]
        hidden_state = self.encoder(hidden_state)

        # chunk
        boundary_prob, boundary_mask, selected_probs = self.chunk_module(hidden_state)

        # main network
        concept = hidden_state[boundary_mask]            # shape [701, 2048]
        concept = self.concept_model(concept)

        # dechunk
        concept_merge = self.dechunk_module(concept, boundary_prob, boundary_mask)

        # decoder
        hidden_state = hidden_state + concept_merge * ste_func(selected_probs)
        hidden_state = self.decoder(hidden_state， concept_merge)   # joint decoding

        logits = self.lm_head(hidden_state)
        return logits
\end{python}

\end{document}